\documentclass{mva_style}
\usepackage{graphicx}
\usepackage{caption}
\usepackage{subcaption}
\usepackage{xcolor}
\usepackage{multirow}
\usepackage{enumitem}
\usepackage{amsmath}
\usepackage{amssymb}
\usepackage{soul}
\usepackage[export]{adjustbox}
\usepackage{booktabs}
\usepackage{comment}
\usepackage{hhline}
\usepackage{float}
\usepackage{authblk}

\finalcopy 

\begin{document}
\makeatletter
\renewcommand\AB@affilsepx{, \protect\Affilfont}
\makeatother
\title{Contrastive Knowledge Distillation for Anomaly Detection in Multi-Illumination/Focus Display Images}

\author[1]{Jihyun Lee\thanks{Denote Equal contribution}}
\newcommand\CoAuthorMark{\footnotemark[\arabic{footnote}]}
\author[1]{Hangil Park\protect\CoAuthorMark}
\author[2]{Yongmin Seo}
\author[1]{Taewon Min}
\author[3]{Joodong Yun}
\author[3]{Jaewon Kim}
\author[1]{Tae-Kyun Kim}
\affil[1]{KAIST}
\affil[2]{Hanyang University}
\affil[3]{Samsung Display}
\maketitle

\section*{\centering Abstract}
\textit{
In this paper, we tackle automatic anomaly detection in multi-illumination and multi-focus display images. The minute defects on the display surface are hard to spot out in RGB images and by a model trained with only normal data. To address this, we propose a novel contrastive learning scheme for knowledge distillation-based anomaly detection. In our framework, Multiresolution Knowledge Distillation (MKD) is adopted as a baseline, which operates by measuring 
feature similarities between the teacher and student networks. Based on MKD, we propose a novel contrastive learning method, namely Multiresolution Contrastive Distillation (MCD), which does not require positive/negative pairs with an anchor but operates by pulling/pushing the distance between the teacher and student features.
Furthermore, we propose the blending module that transforms and aggregate multi-channel information to the three-channel input layer of MCD. Our proposed method significantly outperforms competitive state-of-the-art methods in both AUROC and accuracy metrics on the collected Multi-illumination and Multi-focus display image dataset for Anomaly Detection (MMdAD).}

\begin{figure}[!t]
  \begin{center}
     \includegraphics[width=8cm]{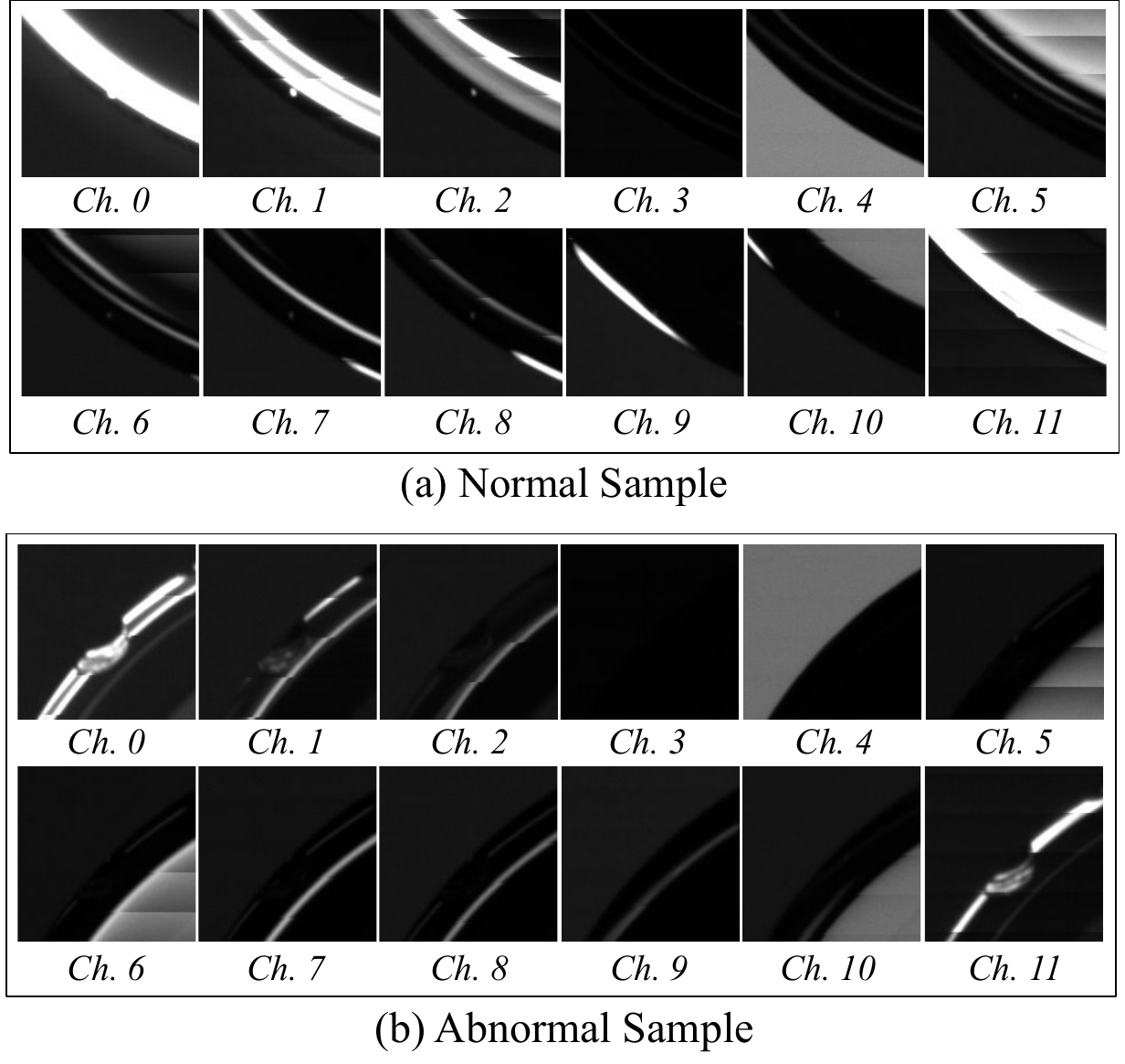}
  \end{center}
  \vspace{-1.2\baselineskip}
  \caption{\textbf{Normal and abnormal samples in the collected dataset (MMdAD).} An image sample consists of 12 channels, each of which is captured with different illumination and focal length settings.}
  \label{fig:data_sample}
  \vspace{-1\baselineskip}
\end{figure}

\section{Introduction}

Anomaly detection is one of the most practical applications in industrial vision, especially for manufacturing. During manufacturing, automatic defect detection is crucial in increasing production efficiency.  
Recent deep learning-based methods 
have shown high accuracies on public anomaly detection benchmarks \cite{rd4ad, dra, mkd}. These existing methods, however, consider only general 3-channel (RGB) images as input in their structure.
The minute defects on the display surface are obscure, and the surface's specular reflection even complicates image patterns. Detecting abnormality in the display images with a single RGB input is difficult, thus we collect a dataset in multi-illumination and multi-focal settings. Also, most existing methods \cite{rd4ad, mkd, opengan, transformaly, patchcore, fakeit, DUIAD, panda, dn2, sspcab, ocrgan} are trained in an unsupervised manner, i.e. using only normal data. Though this is a main setting for anomaly or out-of-distribution detection (also called one-class learning or open set recognition), there has been a need in supervised learning \cite{dra} and semi-supervised learning \cite{ssad, memseg}. We need a mechanism to incorporate a portion of abnormal data, when they are available, to model learning to enhance its accuracy.

This research aims at automatically detecting minute defects on the display surface using the multi-illumination and multi-focal images, and effectively exploiting a small portion of defect data. To achieve this goal, we propose a novel contrastive learning scheme and a blending module for knowledge distillation-based anomaly detection. We use Multiresolution Knowledge Distillation-based method (MKD) \cite{mkd} as our baseline, since it has shown outperforming other competitive methods on industrial defection benchmarks. MKD with the teacher and student network is trained to higher the feature similarities of the two networks over normal data. When abnormal data is seen during inference, the student network diverges from the teacher network, yielding low feature similarities. To extend MKD to supervised learning settings, we propose a novel contrastive learning method, namely Multiresolution Contrastive Distillation (MCD). Whereas conventional contrastive learning requires negative/positive pairs with an anchor, our framework is trained by pulling/pushing the distance between the teacher and student features based on the class (i.e., normal or abnormal) of the input image. MKD in its original architecture receives an RGB three-channel image. We thus additionally propose the blending module, composed of three convolutional layers, which transforms and aggregates multi-channel information to the input layer of MKD exploiting MMdAD information fully. In the experiments by our multi-illumination and multi-focus dataset, the proposed method significantly outperforms competitive state-of-the-art methods in both AUROC and accuracy metrics.

In summary, our contributions are three-fold: (1) We propose a novel contrastive learning method for knowledge distillation-based anomaly detection. (2) We propose the blending module that effectively aggregates multi-channel information to the input of MKD. (3) Our proposed method achieves state-of-the-art anomaly detection results on the MMdAD dataset.
\vspace{-1.75\baselineskip}

\section{Related Work on Anomaly Detection}
Recent deep learning-based methods have shown high accuracies on public anomaly detection benchmarks. Anomaly detection tasks are also highly relevant to out-of-distribution detection, openset recognition, and one-class learning \cite{ocsvm, disaugclr, fcdd} in literature. Most existing methods~\cite{rd4ad, mkd, opengan, transformaly, patchcore, fakeit, DUIAD, panda, dn2, sspcab, ocrgan} are learnt in an unsupervised way i.e. only using normal data. The knowledge distillation-based methods \cite{rd4ad, mkd, ast} have shown the highest accuracies on industrial defect benchmarks, while various other methods \cite{opengan, transformaly, patchcore, fakeit} showed good performance on generic image classification, or weather/medical datasets.  
Since real abnormal data is hard to obtain, synthetic data can be generated by GAN or diffusion models and exploited in \cite{opengan, fakeit}. When a portion of abnormal data is available, a binary classifier can be trained and applied. However, than naively applying a binary classifier, a supervised method designed to anomaly detection \cite{dra} has delivered superior detection accuracies. Similar, anomaly detection with semi-supervised learning was proposed in \cite{ssad, memseg}. 
Contrastive learning has also been applied to 
anomaly detection \cite{driverad, contra_ano, csi, mean-shift, disaugclr, cfa}. 






\begin{figure*}[ht]
\noindent
  \begin{center}
    \includegraphics[width=17.5cm]{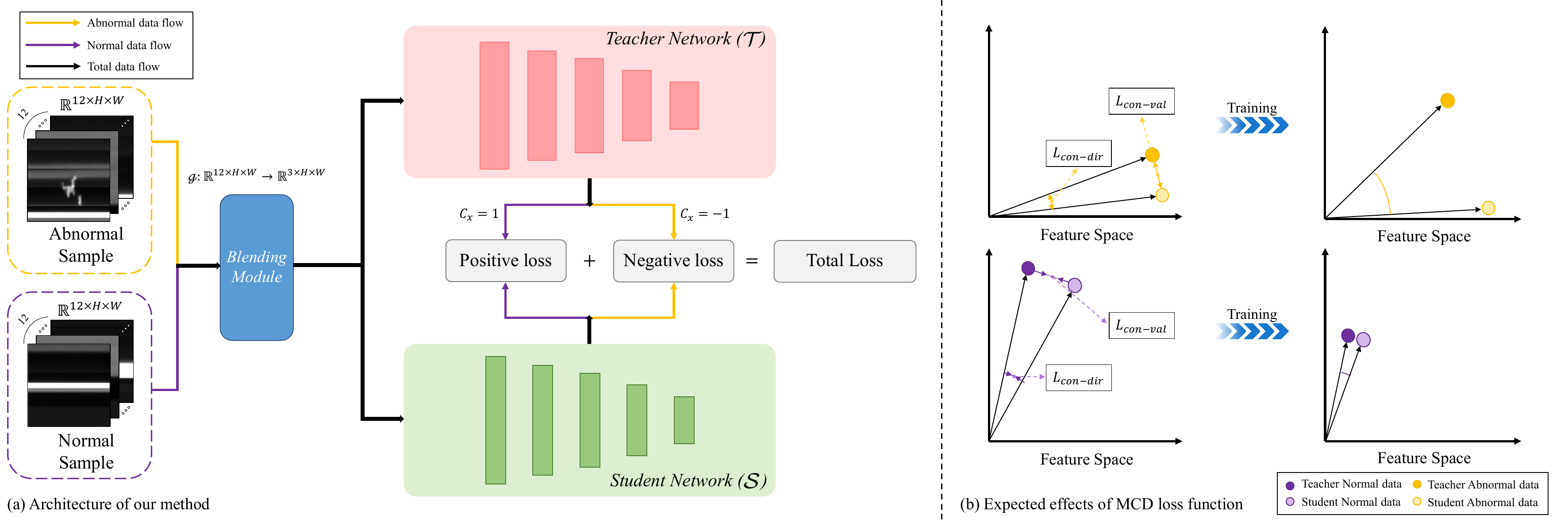} 
  \end{center}
  \vspace{-\baselineskip}
  \caption{\textbf{(a) Method overview.} We propose a novel contrastive learning scheme for knowledge distillation-based anomaly detection. We also introduce the blending module to fully exploit the input multi-illumination/focus channel information. \textbf{(b) Effects of our contrastive learning.} Our contrastive loss pulls/pushes the distance between the teacher and student features based on the class (i.e., normal or abnormal) of the input image.}
  \vspace{-\baselineskip}
  \label{fig:overview}
\end{figure*}

\label{sec:mmdad}
\section{Multi-Illumination and Multi-Focus Display Image Dataset for Anomaly Detection (MMdAD)}

The Multi-Illumination and Multi-Focus Display Image Dataset for Anomaly Detection (MMdAD) consists of 12-channel images, with 134 anomalous and 327 normal display images. Each image has a resolution of 150 x 150 pixels. The channels in these images represent product images in grayscale, varying in illumination and focal length. The dataset includes cropped local regions of display images that have a high likelihood of containing anomalies. The abnormal display data shows a range of defects.
A key distinction between MMdAD and other public anomaly detection benchmarks is that normal samples in MMdAD can also have anomaly-like features. As a result, to differentiate between normal and abnormal images, the MMdAD dataset requires an understanding of the correlation between channels. Figure~\ref{fig:data_sample} provides examples of the samples.

\section{Methodology}
\subsection{Preliminaries}
Multiresolution Knowledge Distillation for Anomaly Detection (MKD) \cite{mkd} is one of the state-of-the-art methods specialized in industrial defect detection. MKD consists of two networks: the teacher and student networks. The teacher network is pre-trained on publicly available large-scale image datasets (e.g. ImageNet\cite{imagenet}). The student network is learnt only using normal data in an unsupervised way, while imitating the feature vectors of the teacher network. The multi-layer knowledge distillation loss is defined as:

\vspace{-0.5\baselineskip}
\begin{equation}
\begin{aligned}
&\mathcal{L}_{total}=\sum_{x\in \mathcal{X}_{\mathit{nor}}}(\lambda \mathcal{L}_{val}(x)+ \mathcal{L}_{dir}(x))&\\
\end{aligned}
\label{eqn:total_loss}
\end{equation}
where $\mathcal{X}_{nor}$ is a set of normal images, $\lambda$ is a balancing parameter, $\mathcal{L}_{val}$ is the mean squared error (MSE) loss, and $\mathcal{L}_{dir}$ is the cosine similarity loss between the features of the teacher ($\mathcal{T}$) and student ($\mathcal{S}$) networks. 
\vspace{-0.5\baselineskip}
\begin{equation}
\begin{aligned}
&\mathcal{L}_{val}(x)=\sum_{i=1}^{N_{\mathit{CP}}}(f_{\mathcal{T}}^{i}(x)-f_{\mathcal{S}}^{i}(x))^{2}&\\
\end{aligned}
\label{eqn:mse_loss}
\end{equation}

where $f^{i}$ is the value of activation in layer $i$ and $N_{\mathit{CP}}$ represents a total number of critical layers for distillation. 
\vspace{-\baselineskip}

\begin{equation}
\small
\begin{aligned}
&\mathcal{L}_{dir}(x)=1-\sum_{i=1}^{N_{\mathit{CP}}}\frac{vec(f_{\mathcal{T}}^{i}(x))^{T}\cdot vec(f_{\mathcal{S}}^{i}(x))}{\left \| vec(f_{\mathcal{T}}^{i}(x))\right \| \left \| vec(f_{\mathcal{S}}^{i}(x))\right \|}&\\
\end{aligned}
\label{eqn:dir_loss}
\end{equation}
where $vec(x)$ is a vectorization function transforming a tensor $x$ into a 1-D vector.
\vspace{-0.5\baselineskip}



\subsection{Multiresolution Contrastive Distillation}


While Multiresolution Knowledge Distillation~[3] is shown to be effective in unsupervised anomaly detection, it does not leverage abnormal examples which our setting assumes to be available during training. To address this, we introduce \emph{Multiresolution Contrastive Distillation (MCD)}, which is a novel contrastive learning method to effectively leverage both positive and negative examples (i.e., normal and abnormal display images) in the knowledge distillation-based framework. Motivated by Eq.~\ref{eqn:total_loss}, we formulate our loss to involve knowledge distillation in \emph{two different metric spaces} (i.e., Euclidean and angular spaces) as follows:

\vspace{-\baselineskip}
\begin{equation}
\small
\begin{aligned}
&\mathcal{L}_{MCD} = \sum_{x\in \mathcal{X}} (\lambda \mathcal{L}_{\mathit{con}-\mathit{val}}(x)+ \mathcal{L}_{\mathit{con}-\mathit{dir}}(x)).\\
\end{aligned}
\label{eqn:mcn_loss}
\end{equation}
\noindent In the above equation, $\mathcal{L}_{\mathit{con}-\mathit{val}}$ denotes contrastive distillation loss in \emph{Euclidean space} (analogous to $\mathcal{L}_{\mathit{val}}$ in Eq.~\ref{eqn:total_loss}) and is defined as:
\vspace{-0.75\baselineskip}
\begin{equation}
\small
\begin{aligned}
&\mathcal{L}_{\mathit{con}-\mathit{val}}(x) = \sum_{i=1}^{N_{\mathit{CP}}}C_x \cdot \mathrm{max} ( \|f_{\mathcal{T}}^{i}(x)-f_{\mathcal{S}}^{i}(x)\|_{2},\, \alpha),\\
\end{aligned}
\label{eqn:mcn_val_loss}
\end{equation}
\vspace{-0.75\baselineskip}

\noindent where $C_x$ denotes the class index of $x$ such that $C_x=1$ if $x$ is positive (i.e., normal), and $C_x=-1$ if $x$ is negative (i.e., abnormal). Note that the above loss enforces the student features to \emph{resemble} teacher features for \emph{normal} image inputs, while it does so to \emph{disresemble} them for \emph{abnormal} image inputs -- through inverting the gradient directions via multiplying the loss by $C_x=-1$. This effectively guides the student network to learn discriminative features for normal and abnormal images with respect to the teacher network. We also found that directly using an MSE-based loss as in Eq.~\ref{eqn:mse_loss} results in training instability due to the unbounded maximum loss magnitude for abnormal examples. Thus, we additionally introduce a $\textrm{max}$ operator and a hyper-parameter $\alpha$ to constrain the upper bound of our loss magnitude to $\alpha$.

In Eq.~\ref{eqn:mcn_loss}, we also introduce contrastive distillation loss in \emph{angular space}, $\mathcal{L}_{\mathit{con}-\mathit{dir}}$, which is defined as:

\vspace{-1\baselineskip}
\begin{equation}
\small
\begin{aligned}
&\mathcal{L}_{\mathit{con}-\mathit{dir}}(x)=1-\sum_{i=1}^{N_{\mathit{CP}}}C_x \cdot \frac{vec(f_{\mathcal{T}}^{\mathit{CP}_{i}}(x))^{T}\cdot vec(f_{\mathcal{S}}^{\mathit{CP}_{i}}(x))}{\left \| vec(f_{\mathcal{T}}^{\mathit{CP}_{i}}(x))\right \| \left \| vec(f_{\mathcal{S}}^{\mathit{CP}_{i}}(x))\right \|}.&\\
\end{aligned}
\label{eqn:mcn_dir_loss}
\end{equation}

\noindent This contrastive loss term is similar to $\mathcal{L}_{\mathit{con}-\mathit{val}}$, except that we consider the feature similarity between teacher and student networks in angular space using cosine similarity (analogous to $\mathcal{L}_{dir}$ in Eq.~\ref{eqn:total_loss}). Using both $\mathcal{L}_{\mathit{con}-\mathit{val}}$ and $\mathcal{L}_{\mathit{con}-\mathit{dir}}$, we can effectively boost our anomaly detection performance by contrasting the teacher and student feature distances between normal and abnormal samples utilizing the dual metric spaces. Also, note that our contrastive distillation does not require a triplet (positive/negative pairs with an anchor) as in a standard contrastive learning method~\cite{triplet}, avoiding training inefficiency caused by triplet sampling.



\subsection{Blending Module}
 
 Even though MKD shows state-of-the-art performance in industrial defect detection, MKD can process only general 3-channel (RGB) images as input. At the same time, the MMdAD dataset consists of 12-channel multi-illumination, multi-focal images. Therefore, we propose the blending module which is a reshaping block to effectively aggregate multi-illumination and multi-focus information in the MMdAD dataset. The blending module first converts 12-channel images to 256-channel latent features that include multi-channel information. The module then reconstructs RGB image using the 256-channel latent features so that MKD can process the reconstructed RGB image. The reconstructed RGB image is generated by selectively containing essential information from every channel for anomaly detection. 
 

\section{Experimental Results}
\vspace{-0.5\baselineskip}

In this section, we conduct experiments to investigate the effectiveness of the proposed method.
\vspace{-0.5\baselineskip}

\begin{description}[leftmargin=0cm]
\item[Dataset and metrics.] Since there is no publicly available multi-illumination/focus image benchmark on anomaly detection, we use the collected Multi-illumination and Multi-focus display image dataset for Anomaly Detection (MMdAD). (Please see Sec.~\ref{sec:mmdad} for dataset details.) We randomly divide the dataset into two splits and conducted 2-fold cross-validation. Each set comprises 124 normal display images and 67 abnormal display images. For evaluation metrics, we use two of the most widely-used evaluation metrics in anomaly detection: Area Under ROC (AUROC) and accuracy.


\noindent \textbf{Implementation details.} During training, we use a batch size of 64, a number of epochs of 200, and $\alpha$ of 1. We use Adam\cite{adam} optimizer for training with a learning rate of 0.001 and momentum parameters of (0.5, 0.999). 

\noindent \textbf{Comparisons with state-of-the-art methods.} In Table 1, we first compare our method with the existing state-of-the-art methods on anomaly detection: One-Class SVM~\cite{ocsvm}, RD4AD~\cite{rd4ad}, DRA~\cite{dra}, and MKD~\cite{mkd}. As the compared methods require 3-channel images as input, we manually choose channels 0, 2, and 6 -- which empirically yield the best accuracy on MKD baseline (see rows 5-10) -- to create input images for the baselines.
In the table, our method outperforms all the compared methods in terms of AUROC by a large margin. Note that ours is shown to achieve better performance than DRA~\cite{dra}, which is also a supervised learning method exploiting abnormal data.

\vspace{-0.5\baselineskip}
\label{table:sota_comparison}
\begin{table}[H]
\caption{\textbf{Comparisons with state-of-the-art methods.} For MKD baseline, we also show the results with varying input channel settings. The best set of input channels for MKD is in bold.}
\vspace{-1.2\baselineskip}
    \begin{center}
        \begin{tabular}{c|c|c}
\hline
Method                         & Input Channels         & AUROC         \\ \hline
One-Class SVM\cite{ocsvm}                  & {[}0, 2, 6{]}          & 0.68          \\ \hline
RD4AD\cite{rd4ad}             & {[}0, 2, 6{]}          & 0.77          \\ \hline
DRA\cite{dra}                 & {[}0, 2, 6{]}          & 0.79          \\ \hline
\multirow{7}{*}{MKD\cite{mkd}} & \textbf{{[}0, 2, 6{]}} & \textbf{0.91} \\  
                             & {[}0, 3, 11{]}         & 0.70          \\ 
                             & {[}1, 3 ,4{]}          & 0.78          \\  
                             & {[}4, 8, 11{]}         & 0.80          \\  
                             & {[}0, 1, 9{]}          & 0.85          \\  
                             & {[}4, 6, 8{]}          & 0.90          \\ \hline
\textbf{Ours}                     & \textbf{{[}0-11{]}}                       & \textbf{0.99}            \\ \hline
\end{tabular}
    \end{center}
\vspace{-1.7\baselineskip}

\end{table}


\item[Ablation study.] We also conduct ablation study to demonstrate the effectiveness of our method. As shown in Table \ref{tabel:ablation}, the blending module makes over 4\% performance improvement in both data splits compared with MKD baseline. Our contrastive learning method also makes further 2-4\% performance improvement in both data splits. Note that we use channels 0, 2, and 6 of the input image for the settings without using the blending module.
In Figure \ref{fig:roc}, we also show ROC curves of the ablation study to further support the effectiveness of the proposed modules. 

\begin{table}[H]
    \caption{\textbf{Quantitative results of ablation study.} BM and CL stand for blending module and contrastive learning, respectively.}
    \vspace{-0.5\baselineskip}
    \label{tabel:ablation}
\begin{tabular}{c|c|c|c|c|c}
\hline
\begin{tabular}[c]{@{}c@{}}Evaluation\\ Metric\end{tabular} & MKD                       & BM                        & CL                        & \begin{tabular}[c]{@{}c@{}}Dataset \\ Split 1\end{tabular} & \begin{tabular}[c]{@{}c@{}}Dataset\\ Split 2\end{tabular} \\ \hline
\multirow{3}{*}{AUROC}                                      & \checkmark &                           &                           & 0.91                                                     & 0.91                                                   \\ \cline{2-6} 
                                                             
                                                            & \checkmark & \checkmark &                           & 0.97                                                     & 0.95                                                    \\ \cline{2-6} 
                                                            & \checkmark & \checkmark & \checkmark & \textbf{0.99}                           & \textbf{0.99}                          \\ \hline
\multirow{3}{*}{Accuracy}                                   & \checkmark &                           &                           & 0.85                                                     & 0.84                                                    \\ \cline{2-6}
                                                            & \checkmark & \checkmark &                           & 0.93                                                     & 0.92                                                    \\ \cline{2-6} 
                                                            & \checkmark & \checkmark & \checkmark & \textbf{0.98}                           & \textbf{0.98}                          \\ \hline

\end{tabular}
\end{table}

\vspace{-\baselineskip}
\begin{figure}[H]
\noindent
  \begin{center}
     \includegraphics[width=0.45\textwidth]{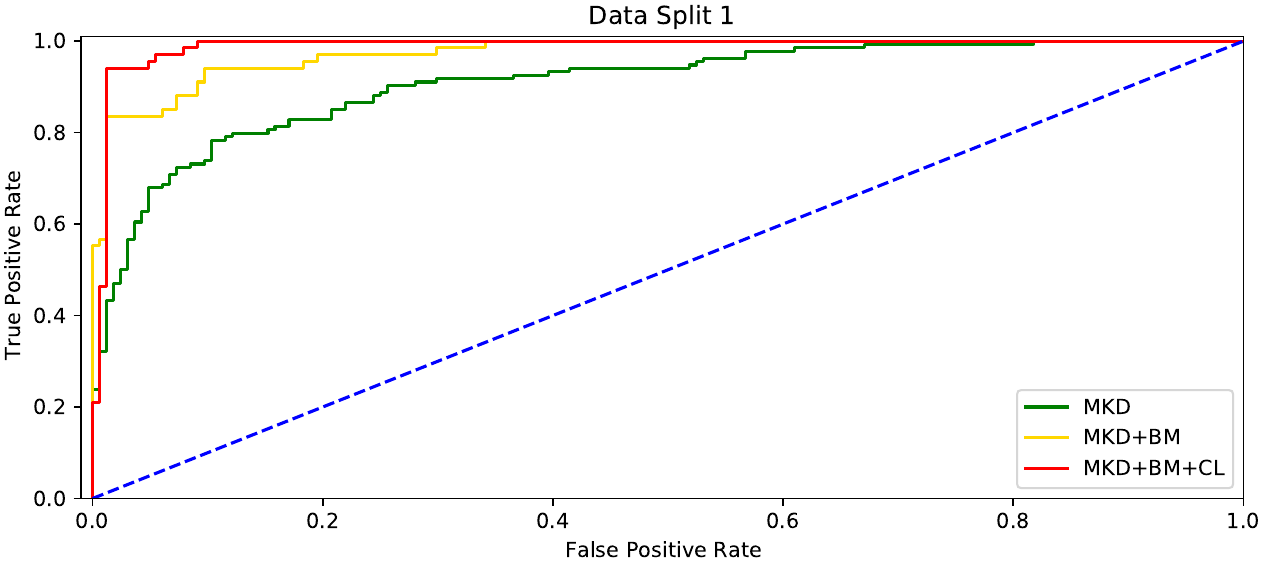}
     
     
  \end{center}
  \vspace{-\baselineskip}
  \caption{\textbf{ROC curves of ablation study.} The curve is obtained from dataset split 1. Our full model (MKD+BM+CL) shows the best performance.}
  \label{fig:roc}
\end{figure}
\vspace{-0.7\baselineskip}

\noindent \textbf{Gradient heatmap visualization.} Figure 4 shows the gradient heatmap images for an abnormal sample. The heatmap images show that our model attends to the abnormal regions while successfully exploiting multi-channel information.

\end{description}

\vspace{-1.5\baselineskip}
\begin{figure}[H]
\noindent
  \begin{center}
    \includegraphics[width=\columnwidth]{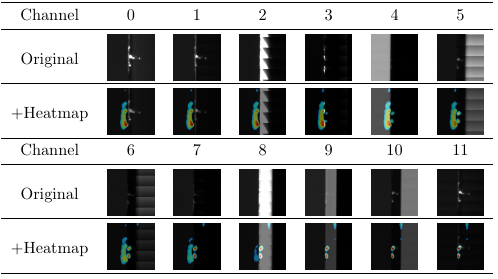} 
  \end{center}
  \vspace{-\baselineskip}
  \caption{\textbf{Gradient heatmaps for an abnormal sample.} Red and blue denotes regions with high and low gradients, respectively.}
  \label{fig:heatmap}
  \vspace{-0.5\baselineskip}
\end{figure}

\section{Conclusion}
In this paper, we propose a novel contrastive learning scheme for knowledge distillation-based anomaly detection on MMdAD. We also introduce the blending module to transform and aggregate multi-channel information to the input layer of our framework.
Our proposed method outperforms competitive state-of-the-art methods on anomaly detection. 

\end{document}